\documentclass{article} 
\usepackage{longcat_style,times}

\usepackage{amsmath,amsfonts,bm}

\def\eqref#1{equation~\ref{#1}}

\def\1{\bm{1}}

\DeclareMathAlphabet{\mathsfit}{\encodingdefault}{\sfdefault}{m}{sl}
\SetMathAlphabet{\mathsfit}{bold}{\encodingdefault}{\sfdefault}{bx}{n}

\usepackage{hyperref}
\usepackage{url}

\usepackage{graphicx}
\usepackage{balance}  
\usepackage{amsmath}
\usepackage{algpseudocode}
\usepackage{latexsym}
\usepackage{multirow}
\usepackage{multicol}
\usepackage{color}
\usepackage{dashrule}
\usepackage{extarrows}
\usepackage{dsfont}
\usepackage{centernot}
\usepackage{hyperref}
\usepackage{natbib}
\usepackage{fancybox}
\usepackage[utf8]{inputenc} 
\usepackage[T1]{fontenc}    
\usepackage{hyperref}       
\usepackage{url}            
\usepackage{amsfonts}       
\usepackage{amsmath}
\usepackage{nicefrac}       
\usepackage{microtype}      
\usepackage{xcolor} 
\usepackage{amsmath}
\usepackage{lipsum}
\usepackage{subcaption}
\usepackage{tabularx}
\usepackage{makecell}
\usepackage{wrapfig}
\usepackage{bm}
\usepackage{multirow}
\usepackage{xcolor} 
\usepackage{bbm}
\usepackage{mdframed}
\usepackage{xspace}
\usepackage{longtable,booktabs}
\usepackage{enumitem}
\usepackage{colortbl}
\usepackage[table]{xcolor}
\usepackage{pifont}
\usepackage[most]{tcolorbox}
\usepackage{xcolor}
\usepackage{fancyvrb}
\usepackage{fvextra}
\definecolor{lightgray}{gray}{0.95}
\usepackage{lineno}
\definecolor{darkblue}{rgb}{0, 0, 0.5}

\hypersetup{colorlinks=true, citecolor=darkblue, linkcolor=darkblue, urlcolor=darkblue}

\usepackage{listings}
\usepackage{xcolor}
\usepackage[ruled,vlined]{algorithm2e}
\usepackage{caption}
\usepackage{float} 
\usepackage{arydshln}
\usepackage{wrapfig} 
\usepackage{tcolorbox}
\tcbuselibrary{theorems}
\tcbuselibrary{most}
\usepackage[dvipsnames]{xcolor}

\usepackage{hyperref}       
\usepackage{url}            
\usepackage{booktabs}       
\usepackage{amsfonts}       
\usepackage{nicefrac}       
\usepackage{microtype}      
\usepackage{xcolor}         
\usepackage{amsmath}
\usepackage{multirow}
\usepackage{multicol}
\usepackage{colortbl}
\usepackage{tcolorbox}
\usepackage{wrapfig}

\usepackage{cleveref}
\usepackage{xspace}

\makeatletter
\DeclareRobustCommand\onedot{\futurelet\@let@token\@onedot}
\def\@onedot{\ifx\@let@token.\else.\null\fi\xspace}

\makeatother

\usepackage{natbib}

\definecolor{light-gray}{gray}{0.6}
\definecolor{front-color}{HTML}{F5FFFA}
\definecolor{Gray}{gray}{0.93}

\definecolor{customTeal}{RGB}{0, 128, 128} 
\definecolor{emphasisColor}{RGB}{255, 0, 0} 
\definecolor{oursBlue}{RGB}{51,202,246}

\definecolor{blue1}{HTML}{508AB2}
\definecolor{green2}{HTML}{BFF6BA}

\newtcbtheorem[]{prompt}{Prompt}%
{colback=SeaGreen!10!CornflowerBlue!10,
 colframe=RoyalPurple!55!Aquamarine!100!,
 fonttitle=\bfseries,
 left=.02in, right=.02in, bottom=.02in, top=.02in,
 before upper={\linespread{1.5}\selectfont}}{prompt}

\renewcommand{\shorttitle}{Uni-Edit: Intelligent Editing Is A General Task For Unified Model Tuning}

\definecolor{darkblue}{rgb}{0, 0, 0.5}
\hypersetup{colorlinks=true, citecolor=darkblue, linkcolor=darkblue, urlcolor=darkblue}

\makeatletter
\renewcommand{\@maketitle}{%
  \vbox{%
    \hsize\textwidth
    \linewidth\hsize
    \vskip -0.5in
    \noindent
    \begin{minipage}{0.99\textwidth}
  \includegraphics[width=0.27\linewidth]{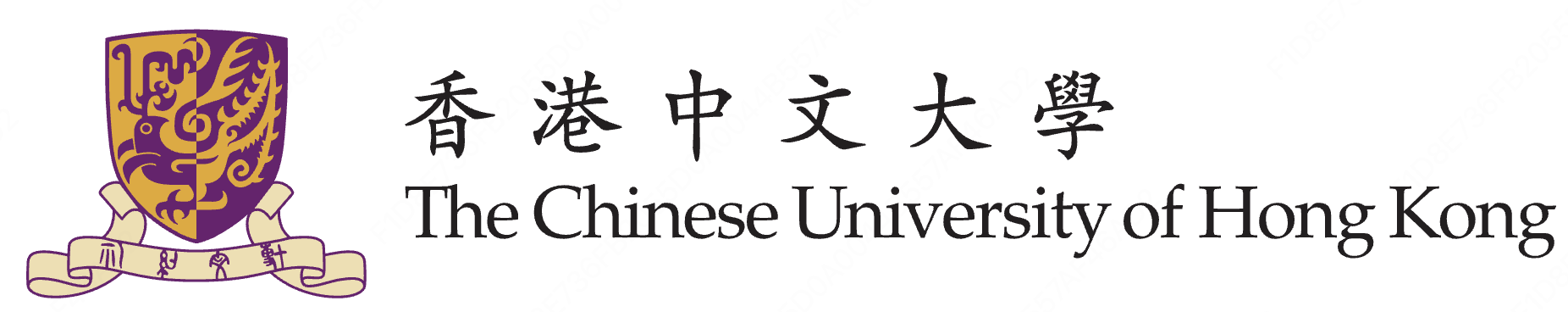}
    \end{minipage}%
    \\
    \rule{\linewidth}{1pt}
    \hspace{0.05\textwidth}%
    \begin{minipage}{0.8\textwidth}
    \end{minipage}

    \centering
    {\LARGE \bfseries\@title\par}
    \vskip 0.1in  
    \def\And{%
      \end{tabular}\hfil\linebreak[0]\hfil%
      \begin{tabular}[t]{c}\bf\rule{\z@}{24\p@}\ignorespaces%
    }
    \def\AND{%
      \end{tabular}\hfil\linebreak[4]\hfil%
      \begin{tabular}[t]{c}\bf\rule{\z@}{24\p@}\ignorespaces%
    }
    \begin{tabular}[t]{c}\bf\rule{\z@}{24\p@}\@author\end{tabular}%
  \vskip 0.05in 
  }
}
\makeatother

\title{Uni-Edit: Intelligent Editing Is A General Task For Unified Model Tuning} 

\makeatletter
\def\@fnsymbol#1{\ensuremath{\ifcase#1\or \dagger\or \ddagger\or
   \mathsection\or \mathparagraph\or \|\or **\or \dagger\dagger
   \or \ddagger\ddagger \else\@ctrerr\fi}}
\makeatother

\newcommand{\homepage}{\raisebox{-1.5pt}{\includegraphics[height=1em]{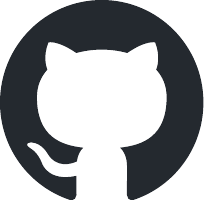}}}
\newcommand{\hfmodel}{\raisebox{-1.5pt}{\includegraphics[height=1em]{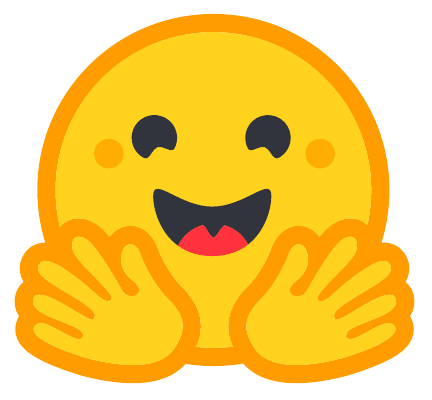}}}

\author{
\begin{tabular}{c}
\textbf{Dian Zheng}$^{1,2}$\textsuperscript{*}
\hspace{0.3em}
\textbf{Manyuan Zhang}$^{2}$\textsuperscript{\dag}
\hspace{0.3em}
\textbf{Hongyu Li}$^{2}$
\hspace{0.3em}
\textbf{Hongbo Liu}$^{3}$
\hspace{0.3em}
\textbf{Kai Zou}$^{4}$
\hspace{0.3em}
\textbf{Kaituo Feng}$^{1}$
\hspace{0.3em}
\textbf{Hongsheng Li}$^{1}$\textsuperscript{\ddag} \\[1ex]
\normalfont $^1$CUHK MMLab 
\quad
\normalfont $^2$Meituan 
\quad
\normalfont $^3$TJU 
\quad
\normalfont $^4$USTC 
\\
{\homepage\ \normalfont 
\texttt{Home: \!\!\!\!\!\url{https://github.com/zhengdian1/Uni-Edit}}} \\
{\hfmodel\ \normalfont \texttt{HF: \!\!\!\url{https://huggingface.co/Uni-Edit}}} \\[0.5ex]
{\normalfont\small 
\textsuperscript{*}Work done while Dian Zheng was an intern at Meituan.
\quad
\textsuperscript{\dag}Project Leader.
\quad
\textsuperscript{\ddag}Corresponding Author.}
\end{tabular}
}

\begin{document}
\maketitle

\begin{figure}[h]
    \centering
    \vspace{-2mm}
    \includegraphics[width=1.0\linewidth]{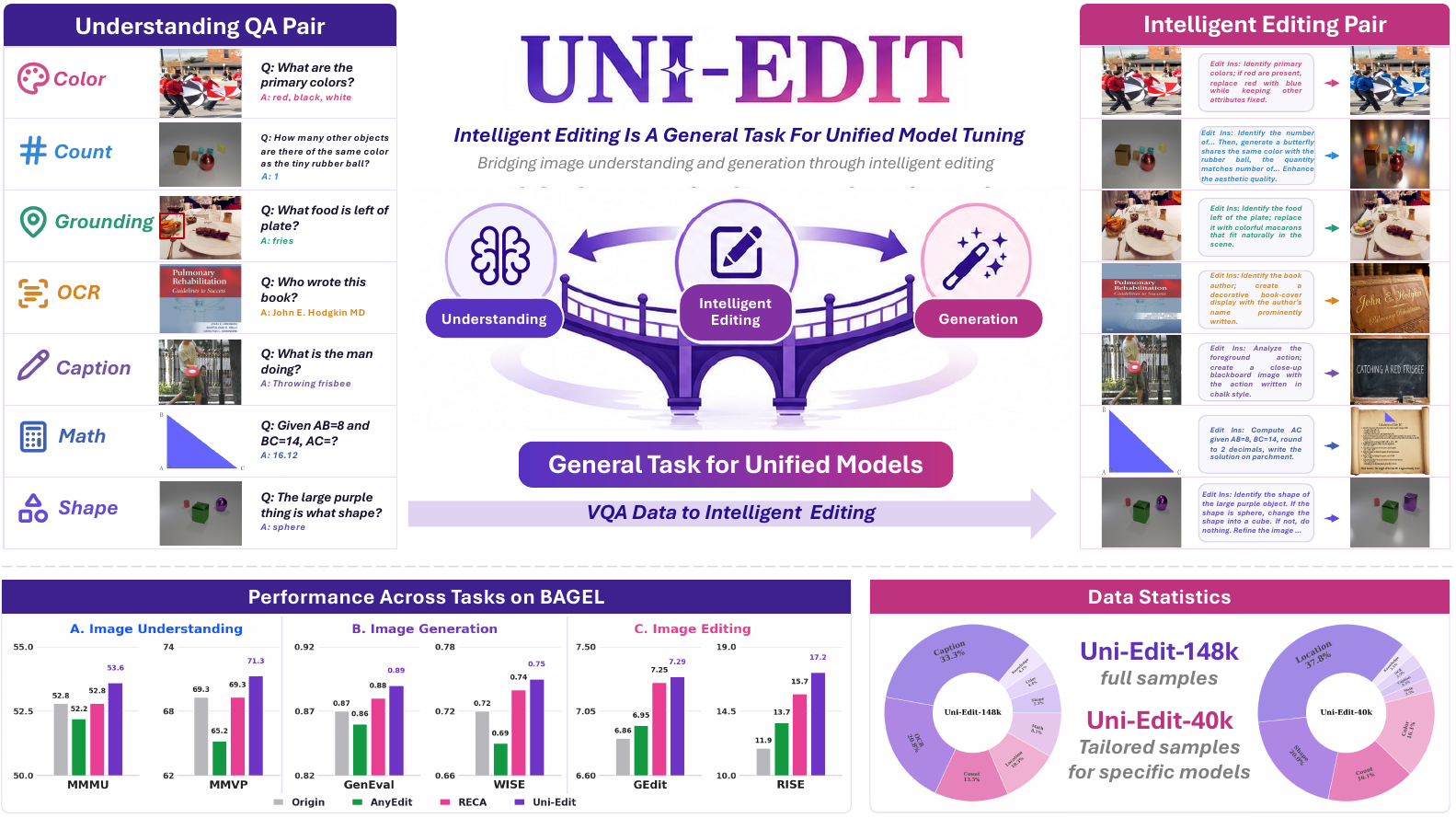}
    \vspace{-6mm}
    \caption{\textbf{Overview of Uni-Edit.} We introduce intelligent image editing as a general tuning task for UMM. By transforming VQA into reasoning-intensive instructions and generating target images via Nano-Pro, we build Uni-Edit-148k. Breaking the trade-offs of existing multi-data mixing strategy, it enhances understanding, generation, and editing using only one task, one dataset, and one training stage. \textbf{Note our automated data construction pipeline can be further scaling if resource permits.}}
    \label{fig:teaser}
\end{figure}

\begin{abstract}
Currently, enhancing Unified Multimodal Models (UMMs) with image understanding, generation, and editing capabilities mainly relies on mixed multi-task training. Due to inherent task conflicts, such strategy requires complex multi-stage pipelines, massive data mixing, and balancing tricks, merely resulting in a performance trade-off rather than true mutual reinforcement. To break this paradigm, we propose Uni-Edit, an intelligent image editing task that serves as the first general task for UMM tuning. Unlike complex mixed pipelines, Uni-Edit improves performance across all three abilities at once using only one task, one training stage, and one dataset.
Specifically, we first identify image editing as an inherently ideal general task, as it naturally demands both visual understanding and generation. However, existing editing data relies on simplistic instructions that severely underutilize a model's understanding capacity. To address this, we introduce the first automated and scalable data synthesis pipeline for intelligent editing, transforming diverse VQA data into complex and effective editing instructions with embedded questions and nested logic. This yields Uni-Edit-148k, pairing diverse reasoning-intensive instructions with high-quality edited images. Extensive experiments on BAGEL and Janus-Pro demonstrate that tuning solely on Uni-Edit achieves comprehensive enhancements across all three capabilities without any auxiliary operations.
\end{abstract}

\section{Introduction}

Unified Multimodal Models (UMMs) aim to integrate general AI capabilities—such as visual understanding, generation, and editing—into a single model, enabling these diverse skills to mutually reinforce one another within a shared latent space. This integration represents a pivotal step toward achieving artificial general intelligence.

While the prospect of UMMs is promising, a fundamental challenge remains: generation and understanding tasks impose conflicting demands on the network architecture. Specifically, in deeper layers, understanding requires high-level semantic information, whereas generation demands fine-grained structural details~\cite{repa}. Consequently, optimizing these tasks within a single model often leads to performance degradation, where enhancing generation impairs understanding and vice versa. Current approaches~\cite{bagel,cui2025emu3_5} primarily address this issue through mixed multi-task training across both pre-training and post-training phases. Due to inherent task conflicts, such a strategy requires complex multi-stage pipelines, massive data mixing, and delicate balancing tricks to strike a performance trade-off rather than true mutual reinforcement. To provide a universally applicable solution for existing UMMs without the prohibitive cost of retraining from scratch, we tackle this conflict during the post-training phase. Instead of modifying model architectures, we focus purely on task, training paradigm and data design to identify \textit{a general tuning task that naturally and simultaneously demands both understanding and generation capabilities.}

To achieve this, we propose Uni-Edit, an intelligent image editing task that serves as the first general task for UMM tuning. Unlike complex mixed pipelines, Uni-Edit simultaneously improves model performance across understanding, generation, and editing using only a single task, a single dataset, and one training stage. Specifically, we first identify image editing as an inherently ideal candidate, as it naturally demands both visual understanding and generation; a deficiency in either capability leads to failure. However, to our knowledge, \textit{no prior work has explored whether editing tuning can enhance understanding performance}. To investigate this, we fine-tune a pre-trained unified model on an existing editing dataset~\cite{yu2025anyedit}. Our results reveal a surprising decline in both understanding and generation capabilities. Upon analysis, we identify the root cause: existing editing data relies on simplistic instructions that severely underutilize the understanding capacity of the model. 

To unlock the true potential of editing, we introduce the first automated and scalable data synthesis pipeline for intelligent editing, transforming diverse VQA data into reasoning-intensive editing instructions. By explicitly embedding questions and nested logic, we force the model to solve underlying problems before performing the edit, thereby balancing the difficulty between understanding and generation during training. Specifically, we source questions from various categories in LLaVA-OneVision-1.5~\cite{ov1_5} and convert them into logically structured editing directives. For example, an instruction might require the model to first solve a mathematical equation and then inscribe both the reasoning process and the final answer onto a blackboard. To generate the corresponding target images, we employ Nano-Banana-Pro (Nano-Pro)~\cite{nano} as our editing backbone. Given the high complexity of our instructions, we supply the original VQA questions and answers as additional context to Nano-Pro to ensure editing accuracy. Finally, we utilize GPT-4o~\cite{gpt4} to rigorously filter the generated data based on instruction following, visual aesthetics, and logical coherence. This complete pipeline ultimately yields Uni-Edit-148k, a curated dataset that pairs complex reasoning instructions with high-quality edited images across a wide range of knowledge domains. 

To validate the efficacy and robustness of Uni-Edit, we use it to fine-tune BAGEL, current state-of-the-art unified model. Our results demonstrate that, even without data balancing or auxiliary modules, Uni-Edit improves BAGEL's performance across generation, editing, and understanding tasks. This finding suggests that with carefully designed editing data, image editing can serve as a general task for unified models tuning. Furthermore, we apply Uni-Edit to both auto-regressive and diffusion-based architectures, revealing distinct advantages of the BAGEL framework. We will fully release our data, models, and code to support the advancement of the UMM community.

We highlight the main contributions of this paper below:
\begin{itemize}
    \item We are the first to propose a general task for unified model tuning, termed intelligent editing, and to investigate how editing tasks affect both understanding and generation capabilities. Through this exploration, we identify that existing editing datasets suffer from overly simplistic and homogeneous instructions.
    \item We propose Uni-Edit-148k, the first editing dataset with intelligent instructions and high-quality images. The instructions cover the broad knowledge required for understanding tasks, and our scalable construction pipeline can be applied to any understanding dataset.
    \item We fine-tune BAGEL, currently the strongest unified model, and achieve simultaneous improvements in generation, understanding, and editing performance without any auxiliary modules. This demonstrates that a general tuning task for unified models indeed exists.
\end{itemize}

\begin{figure}[t]
    \centering
    \includegraphics[width=0.9\linewidth]{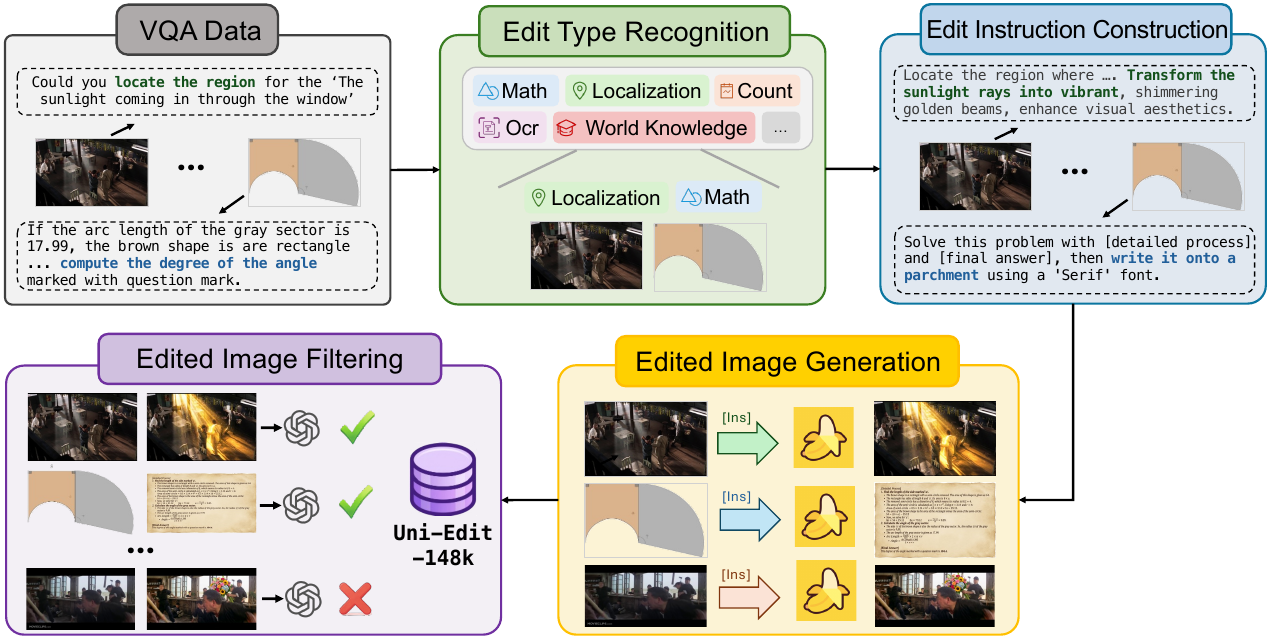}
    \caption{\textbf{Data Construction pipeline}.We first employ GPT-4o to classify the data from LLaVA-OV1.5 into eight distinct edit types, including attribute, caption, math, grounding, and world knowledge. Next, for each category, we use GPT-4o to embed the original question into an editing instruction and explicitly require the model to perform further editing operations based on the answer to the question. This process allows us to construct intelligent editing instructions that encompass general knowledge. Subsequently, we generate the edited images using Nano-Pro and use GPT-4o to filter the data  based on instruction following, visual aesthetics, and coherence.}
    \label{fig:data_curation}
\end{figure}

\section{Related Work}

\subsection{Unified Multimodal Model}

With the maturation of large language models (LLMs) and multimodal LLMs (MLLMs) such as Qwen~\cite{Qwen2.5-VL,yang2025qwen3} and LLaMA~\cite{dubey2024llama, li2024llava}, significant progress has been made in world understanding. Researchers have now shifted their focus toward higher-level capabilities. They aim to equip MLLMs with image generation abilities to construct unified models capable of both understanding and creating the world. 
However, a fundamental challenge has emerged. Understanding and generation tasks inherently conflict as they demand different levels of information granularity in deeper network layers.
To address this, one line of researcher has explored architecture decoupling to mitigate conflicts. Early attempts include fully unified autoregressive models like Emu3~\cite{wang2024emu3}, Chameleon~\cite{team2024chameleon}, VILA-U~\cite{wu2024vila}, LongCat-Next~\cite{longcatnext}. Later work introduced encoder-decoupled architectures such as Janus-Pro~\cite{januspro}. More recent models like BAGEL~\cite{bagel}, HunyuanImage-3.0~\cite{cao2025hunyuanimage} and OneCat~\cite{li2025onecat} abandon pure autoregression. While these approaches have yielded improvements, research like AIA~\cite{aia} indicates that the performance gain is not caused by task conflicts and the conflicts persist.
Another line of work has tackled the issue at the data level. For instance, BAGEL and Emu3.5~\cite{cui2025emu3_5} monitor the convergence speed of understanding, generation, and editing tasks during training. Based on this, they dynamically adjust the data ratio of each task to achieve a balance between understanding and generation performance. Additionally, they train on large-scale interleaved data, enabling the model to perform interleaved image-text generation. However, the instructions in these interleaved data are overly simplistic and monotonous. As a result, they fail to simultaneously improve both understanding and generation performance in a single training phase.
In this work, we seek to identify a task that can universally boost all capabilities of a unified model and find that intelligent editing fulfills this role. Unlike existing approaches, it features complex instructions grounded in diverse world knowledge, making it an ideal candidate of general task for holistic unified training.

\subsection{Image Editing}

Editing has always been a crucial component of image generation. It allows users to personalize existing images and requires models to perform precise, coherent modifications based on instructions. Early generative models~\cite{sd3, podell2023sdxl,flux2024} had limited text understanding capabilities, so instructions were restricted to basic operations like adding, deleting, or changing objects.
Recently, the integration of MLLMs and diffusion heads~\cite{qwenimage, glm_image,gedit,nextstep,editthinker} has yielded impressive results, prompting researchers to explore more complex editing scenarios. For example, AnyEdit~\cite{yu2025anyedit} introduced camera movements, event changes, and world knowledge queries. VQ-VA World~\cite{gou2025vq} further expanded the diversity of world knowledge instructions and improved image quality. \cite{zhuo2025structbench} proposed chart-to-chart editing to test models' understanding of charts.
However, these approaches still fail to cover the full spectrum of general knowledge required for comprehensive understanding tasks. They particularly lack coverage of mathematical reasoning and detailed captioning, which constitute the largest portion of understanding data. In this paper, we propose Uni-Edit and corresponding data construction pipeline that transforms understanding tasks into appropriate editing instructions, achieving broad coverage of general knowledge for the first time. Our experiments demonstrate that, under these conditions, intelligent editing can indeed serve as a general task for unified model tuning.

\begin{table}[t]
    \centering
    \caption{System-level comparison on widely used image understanding benchmarks. Types means tuning BAGEL on different tasks.}
    \resizebox{0.95\columnwidth}{!}{
        \begin{tabular}{lcccccccc}
        \toprule[0.15em]
        \multicolumn{2}{l}{\multirow{2}{*}{\textbf{Dataset}}} & \multirow{2}{*}{\textbf{Types}} & \multicolumn{6}{c}{\textbf{Image Understanding}} \\
        \multicolumn{2}{c} {} & {} & MMMU & MME & MMVet & MathVista & MMVP & MMBench  \\
        \midrule[0.1em]
        \multicolumn{2}{l}{BAGEL data} & Uni. & 52.8 & 2381 & 66.7 & 73.2 & 69.3 & 84.6 \\
        \midrule[0.1em]
        \multicolumn{2}{l}{LLaVA-OV1.5~\cite{ov1_5}} & Und. & 52.6 & 2384 & 65.4 & 73.4 & 68.7 & 84.0 \\
        \multicolumn{2}{l}{Bee~\cite{zhang2025bee}} & Und. & 51.7 & 2317 & 65.0 & 78.3 & 62.7 & 84.5 \\
        \midrule[0.1em]
        \multicolumn{2}{l}{AnyEdit~\cite{yu2025anyedit}} & Edit & 52.2 & 2314 & 63.9 & 71.9 & 65.2 &  84.1 \\
        \bottomrule[0.1em]
    \end{tabular}{}
    }
    \label{tab:data_validation}
\end{table}

\section{Uni-Edit}

In this section, we first identify the inherent difficulty of achieving simultaneous performance gains in understanding, generation, and editing via a single training task—specifically highlighting the high quality of closed-source baseline (BAGEL~\cite{bagel}) data and the coarseness of existing editing datasets. Subsequently, we detail our pipeline for constructing high-complexity editing data designed to bridge this gap. Finally, we introduce our modifications to the BAGEL training paradigm, which are essential for enabling the model to mutually enhance both its understanding and generation capabilities.

\subsection{Identifying the difficulty of the task}

\noindent\textbf{High-quality data used in BAGEL.} We first evaluate the data quality of BAGEL to establish the robustness and difficulty of implementing Uni-Edit on top of it. While tuning with an editing paradigm naturally guarantees performance in editing tasks, its impact on generation and understanding remains uncertain. Given that understanding capability represents the most significant performance gap between current unified models and specialized models, we prioritize validating the quality of BAGEL's understanding data.

To this end, we fine-tuned BAGEL on understanding tasks using two state-of-the-art open-source datasets: Bee~\cite{zhang2025bee} and LLaVA-OV1.5~\cite{ov1_5}. As shown in Table~\ref{tab:data_validation}, these open-source datasets fail to improve BAGEL's performance on general understanding benchmarks; except for Bee, which improves mathematical reasoning. It is important to note that \textit{BAGEL achieves its baseline performance through a unified training paradigm}. The fact that high-quality open-source data—even when \textit{fine-tuned exclusively for understanding}—cannot surpass this baseline demonstrates the exceptional quality of BAGEL's original training data. This finding further underscores the reliability and robustness of our proposed Uni-Edit approach built upon this strong foundation.

\noindent\textbf{Poor quality of existing editing data.} Image editing is inherently a task that challenges both a model's understanding and generation capabilities simultaneously. However, a critical question remains: \textit{can any arbitrary editing dataset effectively boost the performance of a unified model?} To investigate this, we fine-tuned BAGEL using several existing open-source editing datasets and monitored the changes in its understanding performance (\textit{i.e.}, und. ability in unified model serves as the most sensitive indicator for validation). As presented in the Table~\ref{tab:data_validation}, our results reveal a notable decline in understanding performance, even when the editing data incorporates a degree of world knowledge. This suggests that the instructions within current editing datasets are too coarse-grained in terms of semantic understanding to support or enhance the capabilities of a unified model.

\subsection{Intelligent editing data collection}
Transforming understanding tasks into editing assignments presents two primary challenges. 1) given the heterogeneity of understanding responses, which range from multiple-choice and fill-in-the-blanks to captions and chain-of-thought reasoning without explicit annotations, converting each question type into an appropriate editing instruction is non-trivial. 2) since our goal is to simultaneously enhance understanding, generation, and editing capabilities by training solely on editing tasks, ensuring the derived instructions meet this requirement is critical. We address these challenges in detail below and show the complete intelligent edit data curation pipeline in Fig~\ref{fig:data_curation}.

\noindent\textbf{Edit type selection.}
Given the diverse nature of visual understanding, transforming these tasks into editing instructions while retaining their benefit to comprehension requires covering a broad spectrum of understanding categories. This ensures that the model maintains its general capabilities without suffering from catastrophic forgetting of general knowledge. In this work, we categorize understanding tasks into seven distinct domains: chart understanding, mathematical reasoning, OCR, image captioning, attribute recognition (shape, count, and color), spatial reasoning, and world knowledge. Since each understanding sample requires precise instruction conversion, we must first categorize every data point into a specific editing class to facilitate subsequent processing. We select LLaVA-OV1.5 as our source dataset and employ GPT-4o~\cite{gpt4} to classify each entry based on a predefined system prompt (detailed in the Appendix~\ref{eval_prompt}). Based on the model's output, we organize the data into seven distinct categories: \textit{shape}, \textit{color}, \textit{count}, \textit{location}, \textit{OCR}, \textit{caption}, and \textit{math}.

\begin{figure}[t]
    \centering
    \includegraphics[width=1\linewidth]{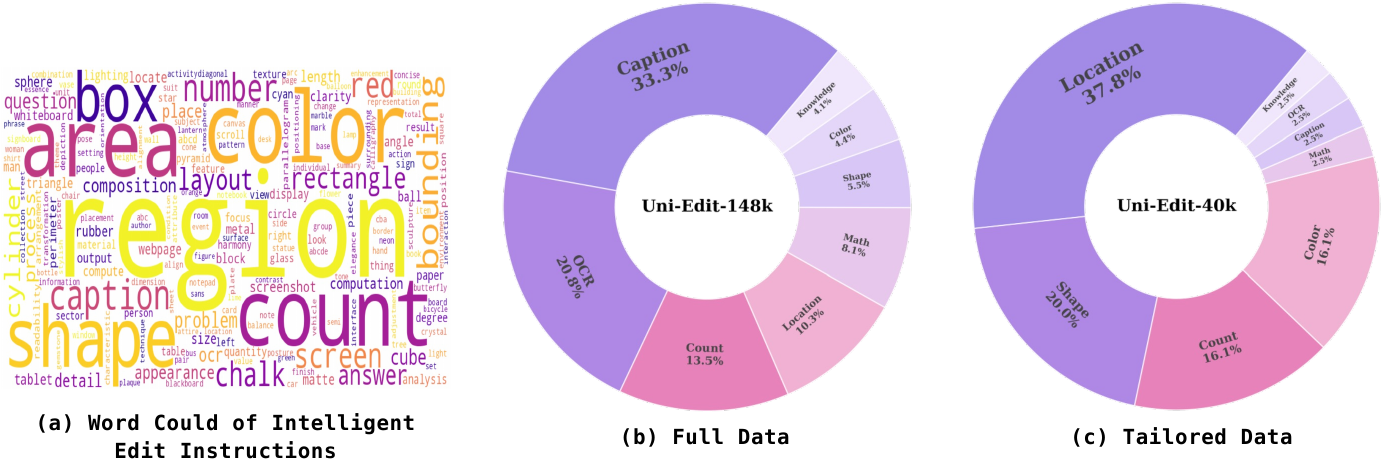}
    \vspace{-4mm}
    \caption{\textbf{Data Distribution} of Uni-Edit-148k and Uni-Edit-40k.}
    \label{fig:distribution}
    \vspace{-4mm}
\end{figure}

\noindent\textbf{Question to edit instruction.}
Based on the assigned category for each understanding sample, we apply targeted transformations to generate specific editing instructions. The full system prompt for transformation is shown in Appendix~\ref{eval_prompt}

$\triangleright$ For \textbf{OCR} and \textbf{caption}, we require the model to first generate a caption or perform OCR on the input image, and then render the derived answer onto a specified writing medium (e.g., a blackboard or letter) using a designated font (e.g., Arial or handwritten style). This approach effectively simulates common requirements in generation and editing workflows.
$\triangleright$ For \textbf{math}, the model is instructed to solve the problem presented in the image and write both the reasoning process and the final answer onto a writing medium, similar to the OCR task. This not only simulates the cognitive process of solving mathematical problems but also enforces precise text generation capabilities.
$\triangleright$ For \textbf{shape}, \textbf{color}, and \textbf{count}, we replicate the full sentence from the understanding question and introduce two variations to enhance diversity. In the first variation, we provide a reference answer and ask the model to verify its correctness, executing different editing instructions based on whether the answer is true or false. In the second variation, we withhold the reference answer and directly instruct the model to add a new object that matches the shape or color of the correct answer. Both methods ensure that the model must correctly solve the underlying understanding problem to execute the edit successfully, while also increasing the variety of editing scenarios.
$\triangleright$ For \textbf{location}, we require the model to localize the object mentioned in the question and replace the object within that specific region with a different item, thereby reinforcing the model's spatial grounding capabilities.
$\triangleright$ For \textbf{knowledge}, given the scarcity of world knowledge samples in LLaVA-OV1.5, we directly incorporate the knowledge subset from AnyEdit~\cite{yu2025anyedit}.

Finally, for all categories except knowledge, we append a directive to enhance the visual aesthetic of the image, thereby serving the dual purpose of improving both generation and editing quality. The examples for each class are shown in Appendix~\ref{type_example}.

\noindent\textbf{Data generation.}
Given the complexity of our editing instructions, we utilize Nano-Pro~\cite{nano}, currently the most powerful image editing model, as our backbone. However, during sampling, we observed that even Nano-Pro frequently makes errors in these complex editing scenarios. To mitigate this, we also provide the original question and answer from the understanding data as context input.

\begin{figure}[t]
    \centering
    \includegraphics[width=0.5\linewidth]{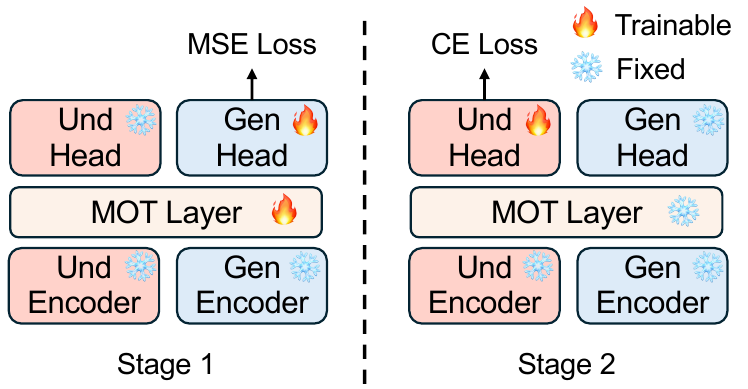}
    \vspace{-2mm}
    \caption{\textbf{Tuning pipeline}. In stage 1, we fine-tune BAGEL on our Uni-Edit data using only the generation loss. In stage 2, we align the distribution of the understanding head with the fine-tuned model using 80k understanding samples. MOT Layer means all of the transformer blocks in BAGEL, Both Und., Gen. heads are a single linear layer.}
    \label{fig:train}
    \vspace{-4mm}
\end{figure}

\noindent\textbf{Data filtering.}
Even with ground-truth context provided, data generated by Nano-Pro still exhibit an error rate of approximately 10\%, we employ GPT-4o~\cite{gpt4} to filter the generated images. Specifically, we provide the model with the original image, the edited image, the editing instruction, and the ground-truth answer from the original understanding dataset. Using a system prompt adapted from VIEScore (Appendix~\ref{eval_prompt}), we instruct the model to evaluate the correctness of the edited image. Based on the filtering rule, we obtain Uni-Edit-148k as shown in Fig~\ref{fig:distribution}. Finally, we conducted ablation studies across all editing data categories to identify which types jointly enhance understanding, generation, and editing, and which might compromise understanding capabilities. Guided by this analysis, we optimized the data sampling proportions. Furthermore, considering that current models like BAGEL still possess limited text-rendering capabilities, tasks heavily reliant on this (OCR, caption and math) are less suitable. Consequently, we retained only a minimal subset of these tasks. This overall refinement process yielded Uni-Edit-40k, a curated dataset tailored to current models. Detailed ablation results are provided in Section~\ref{sec:ablation}, and the final data distribution is shown in Fig~\ref{fig:distribution}.

\subsection{Training Scheme}
We adopt BAGEL as our base model due to its superior understanding capabilities among current unified models and its inherent ability to perform image editing. Validating our Uni-Edit on such a strong baseline ensures the robustness of our conclusions. As shown in Fig~\ref{fig:train}, the full training pipeline comprises two distinct stages, which we describe in the following sections.

\noindent\textbf{Stage 1.} We followed the official BAGEL editing configuration with one key modification: we set the dropout rate for the VAE features (\textit{i.e.}, the generation branch) to 1. This adjustment is designed to compel the model to rely more heavily on the ViT features (\textit{i.e.}, the understanding branch), thereby enhancing understanding performance. This is particularly important because only ViT features are utilized during image understanding tasks. We provide further ablation studies validating this design choice in Sec~\ref{sec:ablation}.

\noindent\textbf{Stage 2.} Since we train the model using only the generation loss, the language modeling head (lm\_head) of the understanding branch remains unoptimized. This can lead to a domain gap between the fine-tuned understanding backbone and the frozen output head. To address this, we employ 80k understanding samples from LLaVA-OV1.5 to fine-tune only the lm\_head, aligning it with the updated parameters of the understanding branch. 
\textbf{Note that this stage is optional, our ablation studies show that Uni-Edit achieves performance gains across all tasks even without this alignment step.}

\begin{table}[t]
    \centering
    \caption{System-level comparison on widely used image understanding, generation and editing benchmarks. $\dagger$ means the benchmark is tested in thinking mode. The row in gray is result of stage 2.}
    \resizebox{1.\columnwidth}{!}{
        \begin{tabular}{llcccccccccc}
        \toprule[0.15em]
        \multicolumn{2}{l}{\multirow{2}{*}{\textbf{Method}}} & \multicolumn{5}{c}{\textbf{Image Understanding}} & 
        \multicolumn{2}{c}{\textbf{Image Generation}} & 
        \multicolumn{3}{c}{\textbf{Image Editing}} \\
        \multicolumn{2}{c} {} & MMMU & MME & MathVista & MMVP & MMBench & GenEval & WISE$\dagger$ & ImgEdit$\dagger$ & RISE$\dagger$ & GEdit$\dagger$ \\
        \midrule[0.1em]
        \multicolumn{2}{l}{BAGEL} & 52.8 & 2381 & 73.2 & 69.3 & 84.6 & 0.87 & 0.72 & 3.28 & 11.9 & 6.86 \\
        \multicolumn{2}{l}{BAGEL-RecA} & 52.8 & 2381 & 73.2 & 69.3 & 84.6 & 0.88 & 0.74 & \textbf{3.75} & 15.7 & 7.25 \\
        \multicolumn{2}{l}{BAGEL-AnyEdit} & 52.2 & 2314 & 71.9 & 65.2 &  84.1 & 0.86 & 0.69 & 3.39 & 14.7 & 6.95 \\
        \midrule[0.1em]
        \multicolumn{2}{l}{\textbf{BAGEL-Uni-Edit (Ours)}} & \textbf{53.6} & \textbf{2405} & \textbf{73.8} & \textbf{71.3} & \textbf{85.5} & \textbf{0.89} & \textbf{0.75} & 3.51 & \textbf{17.2} & \textbf{7.29} \\
        \rowcolor{gray!19} & & 54.2 & 2412 & 74.3 & 72.1 & 86.0 & 0.89 & 0.74 & 3.46 & 16.7 & 7.25 \\
        \midrule[0.1em]
        \multicolumn{2}{l}{a. w/o VAE dropout} & 52.9 & 2344 & 71.7 & 69.3 & 84.1 & 0.87 & 0.72 & 3.34 & 15.9 & 7.06 \\
        \multicolumn{2}{l}{b. w/o joint training} & 52.8 & 2381 & 73.2 & 69.3 & 84.6 & 0.86 & 0.73 & 3.27 & 16.7 & 6.81 \\
        \multicolumn{2}{l}{c. ViT (und.) 224$\times$224} & 52.5 & 2324 & 69.3 & 67.3 & 83.6 & 0.88 & 0.74 & 3.51 & 15.3 & 7.20 \\
        \multicolumn{2}{l}{c. ViT (und.) 224$\times$980} & 51.7 & 2326 & 68.3 & 65.8 & 84.0 & 0.88 & 0.73 & 3.44 & 17.8 & 7.23 \\
        \multicolumn{2}{l}{c. ViT (und.) 378$\times$980} & 51.1 & 2307 & 65.4 & 65.9 & 83.2 & 0.86 & 0.71 & 3.17 & 13.3 & 6.89 \\
        \midrule[0.1em]
        \multicolumn{2}{l}{Janus-Pro} & 41.5 & 1978 & 45.3 & 48.0 & 67.5 & 0.80 &  0.45 & 3.13 & 1.2 & 6.34\\
        \multicolumn{2}{l}{\textbf{+Uni-Edit (Ours)}} & 43.5 & 2067 & 49.2 & 54.3 & 70.4 & 0.82 & 0.52 & 3.43 & 3.4 & 6.87\\
        \bottomrule[0.1em]
    \end{tabular}{}
    }
    \vspace{-1em}
    \label{tab:main}
\end{table}

\section{Experiment}

\subsection{Experimental Setup}

\noindent\textbf{Implementation Details.} We completely follow the training configuration of the official code of BAGEL. Under the FSDP framework, the entire training process for the 14B model took approximately 14 hours on a cluster of 4 nodes, each equipped with 8 NVIDIA H800 (80GB) GPUs.

\noindent\textbf{Evaluation Metrics.} We evaluate our model among image understanding, generation and editing tasks. For understanding, we select 5 widely used benchmarks—MME~\cite{fu2023mme}, MMBench (1.0-EN)~\cite{liu2024mmbench}, MMMU~\cite{yue2024mmmu}, MathVista~\cite{lu2023mathvista}, and MMVP~\cite{mmvp}—which collectively establish a compact yet comprehensive framework for assessing perception, cognition, and multimodal reasoning, with the robust ability to discern the general knowledge embedded in a model. For image generation, we select Geneval~\cite{ghosh2023geneval} and WISE~\cite{niu2025wise} as the benchmarks to evaluate the spatial understanding and reasoning ability of the model; For image editing, we select three widely used benchmarks: ImgEdit~\cite{ye2025imgedit}, GEdit~\cite{gedit}, and RISE~\cite{rise}. The first two evaluate basic editing capabilities, whereas the third focuses on knowledge-based complex editing.

\subsection{Evaluation on the benchmarks}
\textbf{Quantitative Results.} We show the results in Table~\ref{tab:main}, 
fine-tuning with Uni-Edit leads to consistent improvements across understanding, generation, and editing. The gains in editing are expected, given our use of Nano-Pro and an editing-focused training paradigm. However, the simultaneous boost in understanding and generation reveals the significant potential of this training approach.
In terms of generation, the complexity of our editing instructions boosts the understanding and reasoning ability of the model, resulting in substantial gains on the WISE benchmark. Additionally, since GenEval evaluates spatial reasoning and layout, our inclusion of attribute-aware and spatial-aware data naturally enhances performance in these domains.
Regarding understanding, the broad coverage of general knowledge in Uni-Edit drives improvements across multiple benchmarks. Most notably, the performance increases on MMMU and MathVista confirm that the model has strengthened its general knowledge and mathematical reasoning. While methods like RecA~\cite{reca} focus on improving text-image alignment, unfreezing their understanding parameters often degrades their core comprehension capabilities. In contrast, Uni-Edit avoids this trade-off; our intelligent editing mechanism consistently enhances performance across all tasks.

\textbf{Architecture beyond BAGEL.} To validate the generality of Uni-Edit, we further tuned on Janus-Pro, we still boost it in all of the three tasks.

\textbf{Compared with Existing Editing Dataset.} We compare these results against BAGEL fine-tuned on AnyEdit, showing that standard editing data is insufficient. It proves that only a dataset like Uni-Edit, which features intelligent instructions grounded in general knowledge, can effectively serve as a universal task for tuning unified models. 

\begin{figure*}[t]
    \centering
    \includegraphics[width=0.98\linewidth]{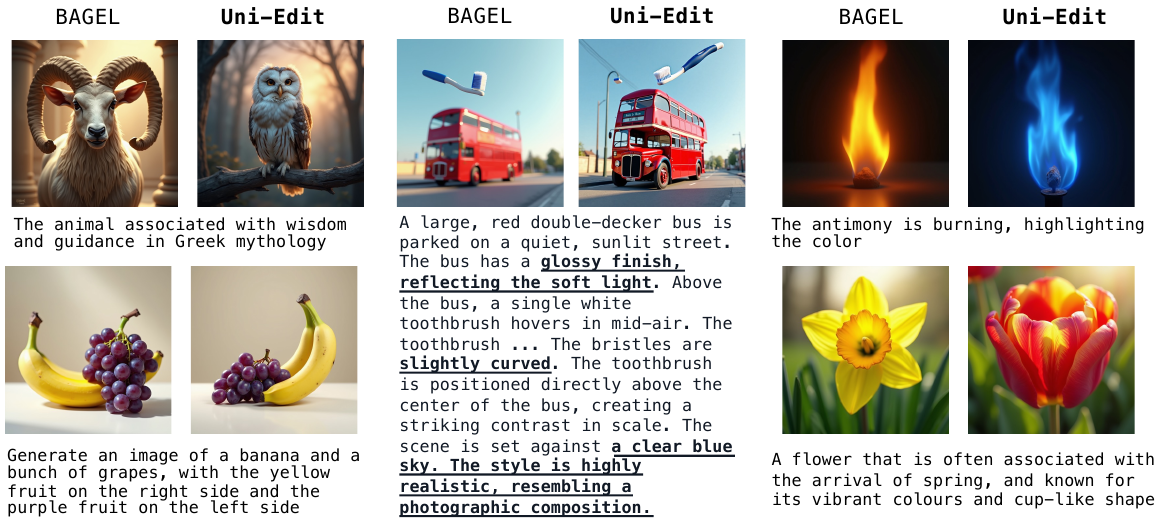}
    \vspace{-2mm}
    \caption{\textbf{Comparison of image generation results between Uni-Edit and BAGEL}. Tuned on Uni-Edit, the model demonstrates significant improvements in prompt understanding, knowledge reasoning, spatial perception, image composition, and aesthetic quality.}
    \label{fig:gen}
    \vspace{-1em}
\end{figure*}

\begin{figure*}[t]
    \centering
    \includegraphics[width=0.98\linewidth]{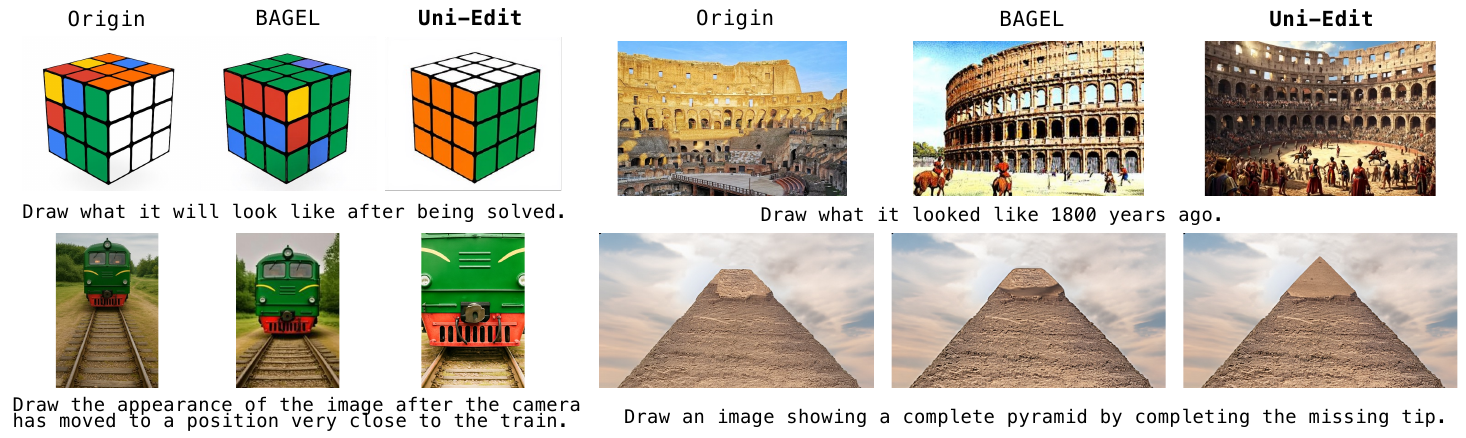}
    \vspace{-2mm}
    \caption{\textbf{Comparison of image editing results between Uni-Edit and BAGEL}. Tuned on Uni-Edit, the model shows significant improvements in instruction following, logic, and spatial reasoning.}
    \label{fig:edit}
    \vspace{-4mm}
\end{figure*}

\textbf{Visualization.} To further validate the effectiveness of Uni-Edit, we show the visulization of image generation and editing compared with BAGEL on Fig~\ref{fig:gen} and Fig~\ref{fig:edit}. The samples involving Rubik's Cube and knowledge reasoning were not included in our training set, demonstrating that Intelligent Editing induces generalized reasoning capabilities.

\subsection{Ablation Study}
\label{sec:ablation}
\noindent\textbf{The effectiveness of VAE feature dropout.} To enable Classifier-Free Guidance (CFG), BAGEL is trained with a 30\% probability of dropping the VAE features. However, our goal is to leverage editing tasks to enhance understanding performance, and understanding tasks do not utilize VAE features. Consequently, we completely discard the VAE input during training. As shown in Table~\ref{tab:main}a, retaining the VAE input leads to performance degradation across all three tasks, with the decline being particularly pronounced in understanding. This suggests that when trained on editing tasks, BAGEL tends to rely more heavily on VAE features for editing, which is detrimental to the improvement of understanding capabilities. Therefore, when employing intelligent editing as a general task, it is crucial to rely solely on ViT features to enhance network consistency.

\noindent\textbf{The function of each type of data.}
Here, we explore the impact of each editing category on the three tasks of the unified model. To ensure fairness, we use 6,000 samples for each category. As shown in Table~\ref{tab:data_ratio}, attribute data (color, shape, count), which is most visually intuitive for generation, yields significant performance gains in generation and editing tasks. However, due to its limited coverage of understanding-related knowledge, it leads to a decline in understanding performance.
Math data improves performance on math-related benchmarks like MathVista. Conversely, converting math, caption, and OCR tasks into editing instructions involves writing answers on a blackboard or letter. This format provides little benefit to general generation and editing tasks and can even have a negative impact. This may also stem from BAGEL's limited capability in generating precise, large-scale text.
Location data significantly boosts performance on GenEval and WISE, as both benchmarks involve spatial reasoning. This confirms that editing tasks can indeed enhance generation performance. Knowledge data promotes performance on multidisciplinary benchmarks like MMMU and knowledge-related generation benchmarks like WISE, demonstrating that editing instructions containing general knowledge can simultaneously foster both generation and understanding.

Based on this fine-grained experimental comparison, we adjusted the data ratios for different categories and ultimately achieved the optimal overall performance.

\noindent\textbf{Impact of joint training.} Since Uni-Edit employs an editing-based training paradigm, a natural question arises: \textit{would freezing BAGEL's understanding branch lead to greater performance gains in the generation branch?} We present the results of this investigation in Table~\ref{tab:main}b. Surprisingly, freezing the understanding branch yields almost no improvement in the model's generation and editing capabilities. This indicates that within the BAGEL architecture, the understanding branch plays an auxiliary role in supporting the generation branch, effectively meaning \textit{understanding aids generation}. At the same time, training with the generative paradigm inversely enhances the model's understanding performance. This highlights the importance of complex editing instructions that cover broad general knowledge and validates the true effectiveness of intelligent editing.

\begin{table}[t]
    \centering
    \caption{System-level comparison of different edit type.}
    \resizebox{0.8\columnwidth}{!}{
        \begin{tabular}{lccccccc}
        \toprule[0.15em]
        \textbf{Dataset} & \textbf{MMMU} & \textbf{MathVista} & \textbf{GenEval} &\textbf{WISE} & \textbf{RISE} \\
        \midrule[0.1em]
        BAGEL (origin) &  52.8 & 73.2 & 0.87 & 0.72 & 11.9   \\
        Uni-Edit (stage 1) &  53.6 & 73.8 & 0.89 & 0.75 & 17.2   \\
        \midrule[0.1em]
        Attibute &  52.4 & 73.4 & 0.88 & 0.74 & 18.6   \\
        Location &  53.2 & 74.2 & 0.89 & 0.75 & 15.6  \\
        Math & 52.9 & 74.4 & 0.86 & 0.73 & 15.3  \\
        Caption\&OCR & 52.1 & 72.2 & 0.86 & 0.73 & 16.9  \\
        Knowledge &  54.2 & 72.6 & 0.87 & 0.75 & 17.8   \\
        \bottomrule[0.1em]
    \end{tabular}{}
    }
    \vspace{-1em}
    \label{tab:data_ratio}
\end{table}

\noindent\textbf{Impact of alignment stage.}
Here, we explore the impact of the alignment stage on both the understanding and generation branches. Note that since certain generation and editing benchmarks require text reasoning, the language modeling head (lm$\_$head) is utilized. The results are presented in the gray line of Table~\ref{tab:main}. It can be observed that aligning the LM head with the original model distribution indeed leads to certain improvements in understanding tasks. However, this alignment results in a declining trend for generation and editing performance. This suggests that for the understanding branch, there is a gap between recognizing a concept in the latent space during editing and explicitly outputting it during understanding tasks. 

\noindent\textbf{Impact of source image resolution.} 
Here we tested the influence of input resolution to UMM tuning. As shown in Table~\ref{tab:main}c, the original resolution of 224 $\times$ 518 yields the best results. Whether we increase the resolution limit to 980, which matches the understanding task, or decrease it to 224, the model's performance drops significantly. Aligning the resolution completely with the understanding task at 378 $\times$ 980 also results in poor performance, with the most noticeable decline in understanding capabilities. This indicates that under unified paradigm training, reducing the image resolution severely impacts the model's understanding ability. Conversely, increasing the resolution may increase the learning difficulty because BAGEL was not originally trained on such large resolutions for editing tasks. In short, adjusting the ViT input resolution during unified tuning does not bring performance gains.

\section{Conclusion and Limitations}

In this work, we propose intelligent editing as a general task for UMM tuning. We introduce Uni-Edit-148k, a high-quality dataset that transforms diverse understanding queries into complex editing instructions, effectively bridging the semantic gap that limits existing editing data. Our experiments on the BAGEL framework demonstrate that training with Uni-Edit enables the mutual reinforcement of capabilities, yielding simultaneous improvements in understanding, generation, and editing without the need for intricate data balancing strategies. These findings not only validate the efficacy of intelligent editing as a general task for unified modeling but also offer a scalable pathway toward developing more robust and versatile multimodal systems.

\textbf{Limitations.} The effectiveness of our data is partially bottlenecked by the base model's capabilities. For example, the Caption, Math, and OCR data in Uni-Edit-148k are not fully utilized in BAGEL and Janus-Pro tuning due to their limited text rendering performance.

{
    \small
    \bibliographystyle{unsrt}
    \bibliography{conference}
}

\clearpage

\appendix

\section{System Prompt}
\label{eval_prompt}

\begin{tcolorbox}[
    colback=lightgray!10,
    colframe=black,
    title={\textbf{Task Type Classification System Prompt}},
    breakable
]
\vspace{0.5em}
\begin{Verbatim}[breaklines=true, breaksymbol={}, fontsize=\tiny]
    "You are an expert data processor. Your task is to analyze the input data ('original question', 'original answer', 'image_path') and produce a structured JSON output that includes the original data plus two new fields: `task_category and process_answer`.\n\n"
    "Your output must be a single JSON object.\n\n"
    "**1. Determine the `task_category`:**\n"
    "Analyze the 'original question' and classify it into one of the following categories:\n"
    "- `shape`: For questions about object shapes (e.g., 'What shape...?').\n"
    "- `count`: For questions about quantity (e.g., 'How many...?').\n"
    "- `bool`: For questions that can be answered with 'yes' or 'no' (e.g., 'Is there...?').\n"
    "- `color`: For questions about color (e.g., 'What color...?').\n"
    "- `location`: For questions about spatial position or spatial relationship (e.g., 'Where is...?').\n"
    "- `caption`: For requests to describe an image.\n"
    "- `ocr`: For requests to only describe the text in the image, which is different from `caption`.\n"
    "- `math`: For mathematical problems.\n"
    "- `multi-choice`: For multi-choice problems.\n"
    "- `others`: For problems that not contain in the types above (e.g., 'What time...?', ''What does the traffic light looks like...?'').\n\n"
    "**2. Extract the `process_answer`:**\n"
    "- **For VQA types** (`shape, count, bool`, etc.): Extract only the core, direct answer from the 'original answer'. Ignore reasoning or filler text.\n"
    "- **For `caption, multi-choice, math and others`**: The `process_answer` should be an exact copy of the 'original answer'.\n\n"
    "**3. Construct the Final JSON Output:**\n"
    "Combine the original input fields with your generated fields into a single JSON object. The keys should be: `task_category, original_question, original_answer, process_answer, image_path`.\n"
    "Provide only the final JSON object and nothing else.\n\n"
    "--- Example 1: VQA (Shape) ---\n"
    "Input:\n"
    "original question: What shape is the large object in front of the small shiny object right of the big matte sphere?\n"
    "original answer: According to the shape and spatial relationship of the existing objects in the image, the answer is sphere\n"
    "image_path: temp.png\n\n"
    "Output:\n"
    "{\n"
    '  "task_category": "shape",\n'
    '  "original_question": "What shape is the large object in front of the small shiny object right of the big matte sphere?",\n'
    '  "original_answer": "According to the shape and spatial relationship of the existing objects in the image, the answer is sphere",\n'
    '  "process_answer": "sphere",\n'
    '  "image_path": "temp.png"\n'
    "}\n"
    "--- Example 2: VQA (Yes/No) ---\n"
    "Input:\n"
    "original question: Is the sky blue in the picture?\n"
    "original answer: Based on the visual evidence, the sky in the provided image is clearly blue. So yes\n"
    "image_path: sky.jpg\n\n"
    "Output:\n"
    "{\n"
    '  "task_category": "bool",\n'
    '  "original_question": "Is the sky blue in the picture?",\n'
    '  "original_answer": "Based on the visual evidence, the sky in the provided image is clearly blue. So yes",\n'
    '  "process_answer": "yes",\n'
    '  "image_path": "sky.jpg"\n'
    "}\n"
    "--- Example 3: Captioning ---\n"
    "Input:\n"
    "original question: Provide a brief description of the image.\n"
    "original answer: A golden retriever is playing fetch in a park on a sunny day.\n"
    "image_path: dog.png\n\n"
    "Output:\n"
    "{\n"
    '  "task_category": "caption",\n'
    '  "original_question": "Provide a brief description of the image.",\n'
    '  "original_answer": "A golden retriever is playing fetch in a park on a sunny day.",\n'
    '  "process_answer": "A golden retriever is playing fetch in a park on a sunny day.",\n'
    '  "image_path": "dog.png"\n'
    "}\n"
    "--- Example 4: VQA (Count) ---\n"
    "Input:\n"
    "original question: How many cars are in the parking lot?\n"
    "original answer: After careful counting, I can confirm there are 5 cars.\n"
    "image_path: parking.jpg\n\n"
    "Output:\n"
    "{\n"
    '  "task_category": "count",\n'
    '  "original_question": "How many cars are in the parking lot?",\n'
    '  "original_answer": "After careful counting, I can confirm there are 5 cars.",\n'
    '  "process_answer": "5",\n'
    '  "image_path": "parking.jpg"\n'
    "}\n"
    "--- Example 5: Math ---\n"
    "Input:\n"
    "original question: If the area of the lime parallelogram is 48, the area of the gray sector is 25.12 and the angle $\\theta$ is vertical to $\\delta$, compute the length of the side of the lime parallelogram marked with question mark. Assume $\\pi=3.14$. Round computations to 2 decimal places.\n"
    "original answer: The length of the hypotenuse of the blue triangle is 24 and the length of the side opposite to the degree of the angle marked with \"$\\delta$\" is 13, so the degree of the angle marked with \"$\\delta$\" equals $\\arcsin(\\frac{13}{24}) = \\arcsin(0.54) = 32.68$. The angle $\\theta$ is vertical to the angle $\\delta$ so the degree of the $\\theta$ angle = 32.68. The angle of the gray sector is 32.68 and the area is 25.12 so the radius marked with \"$a$\" can be computed as $\\sqrt{\\frac{25.12}{\\frac{32.68}{360} * \\pi}} = \\sqrt{\\frac{25.12}{0.09 * \\pi}} = \\sqrt{\\frac{25.12}{0.28}} = \\sqrt{89.71} = 9.47$. The length of one of the sides of the lime parallelogram is 9.47, the area is 48 and the angle is 35. So, the sine of the angle is $\\sin(35) = 0.57$, so the length of the side marked with \"?\" is $\\frac{48}{9.47 * 0.57} = \\frac{48}{5.4} = 8.89$. Therefore the final answer is 8.89.\n"
    "image_path: math.jpg\n\n"
    "Output:\n"
    "{\n"
    '  "task_category": "math",\n'
    '  "original question: "If the area of the lime parallelogram is 48, the area of the gray sector is 25.12 and the angle $\\theta$ is vertical to $\\delta$, compute the length of the side of the lime parallelogram marked with question mark. Assume $\\pi=3.14$. Round computations to 2 decimal places.",\n'
    '  "original_answer": "The length of the hypotenuse of the blue triangle is 24 and the length of the side opposite to the degree of the angle marked with \"$\\delta$\" is 13, so the degree of the angle marked with \"$\\delta$\" equals $\\arcsin(\\frac{13}{24}) = \\arcsin(0.54) = 32.68$. The angle $\\theta$ is vertical to the angle $\\delta$ so the degree of the $\\theta$ angle = 32.68. The angle of the gray sector is 32.68 and the area is 25.12 so the radius marked with \"$a$\" can be computed as $\\sqrt{\\frac{25.12}{\\frac{32.68}{360} * \\pi}} = \\sqrt{\\frac{25.12}{0.09 * \\pi}} = \\sqrt{\\frac{25.12}{0.28}} = \\sqrt{89.71} = 9.47$. The length of one of the sides of the lime parallelogram is 9.47, the area is 48 and the angle is 35. So, the sine of the angle is $\\sin(35) = 0.57$, so the length of the side marked with \"?\" is $\\frac{48}{9.47 * 0.57} = \\frac{48}{5.4} = 8.89$. Therefore the final answer is 8.89.",\n'
    '  "process_answer": "The length of the hypotenuse of the blue triangle is 24 and the length of the side opposite to the degree of the angle marked with \"$\\delta$\" is 13, so the degree of the angle marked with \"$\\delta$\" equals $\\arcsin(\\frac{13}{24}) = \\arcsin(0.54) = 32.68$. The angle $\\theta$ is vertical to the angle $\\delta$ so the degree of the $\\theta$ angle = 32.68. The angle of the gray sector is 32.68 and the area is 25.12 so the radius marked with \"$a$\" can be computed as $\\sqrt{\\frac{25.12}{\\frac{32.68}{360} * \\pi}} = \\sqrt{\\frac{25.12}{0.09 * \\pi}} = \\sqrt{\\frac{25.12}{0.28}} = \\sqrt{89.71} = 9.47$. The length of one of the sides of the lime parallelogram is 9.47, the area is 48 and the angle is 35. So, the sine of the angle is $\\sin(35) = 0.57$, so the length of the side marked with \"?\" is $\\frac{48}{9.47 * 0.57} = \\frac{48}{5.4} = 8.89$. Therefore the final answer is 8.89.",\n'
    '  "image_path": "math.jpg"\n'
    "}\n"
\end{Verbatim}
\end{tcolorbox}

\begin{tcolorbox}[
    colback=lightgray!10,
    colframe=black,
    title={\textbf{QA-to-Edit Instruction System Prompt}},
    breakable
]
\vspace{0.5em}
\begin{Verbatim}[breaklines=true, breaksymbol={}, fontsize=\tiny]
### Attribute Bool type
"You are an expert in Multi-Modal Instruction Generation. Your task is to analyze the input data (`original_question`, `process_answer`, `task_category`) and generate a high-quality **`edit_instruction`**.\n\n"
    
    "Your output must be a single JSON object containing the `edit_instruction` field.\n\n"
    
    "**Core Objective:**\n"
    "Create an instruction that requires the editing model to **understand** the visual content (solve the question) to perform the **generation**, while enhancing aesthetic quality.\n\n"
    
    "**Principles:**\n"
    "1. **Relevance:** The edit must logically follow the question.\n"
    "2. **No Answer Leakage:** Do NOT state the answer directly. Instruct the model to **derive** it from the original image.\n"
    "3. **Aesthetics:** End every instruction with a directive to improve visual quality.\n"
    "4. Output the JSON string directly. Do not include Markdown formatting (such as ```json)..\n\n"
    
    "**Category Guidelines:**\n"
    "- **`shape`/`color`**: Use conditional branches in the ORIGINAL scene: 'Identify the [shape/color of the target objects]. Check if the answer is [process_answer]. If yes, [change this attribute into a different one while maintaining other attributes fixed]. If no, [do not modify the objects].'\n"
    
    "**Few-Shot Examples:**\n\n"
    
    "Input: {task_category: 'shape', original_question: 'There is a cyan thing that is the same size as the blue rubber ball; what shape is it?', process_answer: 'cube'}\n"
    "Output: {\"edit_instruction\": \"Identify the shape of the cyan object that is the same size as the blue rubber ball in the original scene. If the shape is cube, changing the shape into clinder. If not, Do NOT change the object. Refine the image with visual appealing effect.\"}\n\n"

    "Input: {task_category: 'color', original_question: 'There is another tiny sphere that is the same material as the small green sphere; what is its color?', process_answer: 'purple'}\n"
    "Output: {\"edit_instruction\": \"Identify the color of the tiny sphere that is the same material as the small green sphere. If the color is purple, changing the color into red. If not, Do NOT change the object. Refine the image with visual appealing effect.\"}\n\n"

### Attribute Generation Type
    "You are an expert in Multi-Modal Instruction Generation. Your task is to analyze the input data (`original_question`, `process_answer`, `task_category`) and generate a high-quality **`edit_instruction`**.\n\n"
    
    "Your output must be a single JSON object containing the `edit_instruction` field.\n\n"
    
    "**Core Objective:**\n"
    "Create an instruction that requires the editing model to **understand** the visual content (solve the question) to perform the **generation**, while enhancing aesthetic quality.\n\n"
    
    "**Principles:**\n"
    "1. **Relevance:** The edit must logically follow the question.\n"
    "2. **No Answer Leakage:** Do NOT state the answer directly. Instruct the model to **derive** it from the original image.\n"
    "3. **Aesthetics:** End every instruction with a directive to improve visual quality.\n"
    "4. Output the JSON string directly. Do not include Markdown formatting (such as ```json)..\n\n"
    
    "**Category Guidelines:**\n"
    "- **`shape`/`color`**: Synthesize elements that are semantically consistent with the `original_question` and `process_answer`, but belong to a different object class than the target in the `original_question`. Ensure the object is definite and specific, where the asked attributes in the `original_question` should be maintained.\n"
    "- ** content format**: Identify the shape/color of the ... Adding ... into the original scene. Refine the image with visual appealing effect"
    
    "**Few-Shot Examples:**\n\n"
    
    "Input: {task_category: 'shape', original_question: 'There is a cyan thing that is the same size as the blue rubber ball; what shape is it?', process_answer: 'cube'}\n"
    "Output: {\"edit_instruction\": \"Identify the shape of the cyan object that is the same size as the blue rubber ball in the original scene. Adding a ring with the same shape of the cyan object into the original scene. Refine the image with visual appealing effect.\"}\n\n"

    "Input: {task_category: 'color', original_question: 'There is another tiny sphere that is the same material as the small green sphere; what is its color?', process_answer: 'purple'}\n"
    "Output: {\"edit_instruction\": \"Identify the color of the tiny sphere that is the same material as the small green sphere. Adding a ball with the same color of the tiny sphere into the original scene. Refine the image with visual appealing effect.\"}\n\n"

### Count Bool Type
"You are an expert in Multi-Modal Instruction Generation. Your task is to analyze the input data (`original_question`, `process_answer`, `task_category`) and generate a high-quality **`edit_instruction`**.\n\n"
    
    "Your output must be a single JSON object containing the `edit_instruction` field.\n\n"
    
    "**Core Objective:**\n"
    "Create an instruction that requires the editing model to **understand** the visual content (solve the question) to perform the **generation**, while enhancing aesthetic quality.\n\n"
    
    "**Principles:**\n"
    "1. **Relevance:** The edit must logically follow the question.\n"
    "2. **No Answer Leakage:** Do NOT state the answer directly. Instruct the model to **derive** it from the original image.\n"
    "3. **Aesthetics:** End every instruction with a directive to improve visual quality.\n"
    "4. Output the JSON string directly. Do not include Markdown formatting (such as ```json)..\n\n"
    
    "**Category Guidelines:**\n"
    "- **`count`**: Use conditional branches in the ORIGINAL scene: 'Identify the [target objects/logic]. Check if the count is [process_answer]. If yes, [perform the action implied by the question, e.g., remove objects or change them]. If no, [do not change/alternative action the number of the objects].'\n"
    
    "**Few-Shot Examples:**\n\n"
    
    "Input: {task_category: 'count', original_question: 'Hint: Please answer the question and provide the final answer at the end.\nQuestion: How many objects are left if you remove all spheres and cylinders?', process_answer: '2'}\n"
    "Output: {\"edit_instruction\": \"Identify the objects number after removing all spheres and cylinders, if the number is 2, remove 2 objects in the image, if not, do not change the number of the object in the image. Refine the image to enhance its aesthetic appeal with vibrant and harmonious visuals.\"}\n\n"

# Count Generation Type
"You are an expert in Multi-Modal Instruction Generation. Your task is to analyze the input data (`original_question`, `process_answer`, `task_category`) and generate a high-quality **`edit_instruction`**.\n\n"
    
    "Your output must be a single JSON object containing the `edit_instruction` field.\n\n"
    
    "**Core Objective:**\n"
    "Create an instruction that requires the editing model to **understand** the visual content (solve the question) to perform the **generation**, while enhancing aesthetic quality.\n\n"
    
    "**Principles:**\n"
    "1. **Relevance:** The edit must logically follow the question.\n"
    "2. **No Answer Leakage:** Do NOT state the answer directly. Instruct the model to **derive** it from the original image.\n"
    "3. **Aesthetics:** End every instruction with a directive to improve visual quality.\n"
    "4. Output the JSON string directly. Do not include Markdown formatting (such as ```json)..\n\n"
    
    "**Category Guidelines:**\n"
    "- **`count`**: Synthesize elements that are semantically consistent with the `original_question` and `process_answer`, but belong to a different object class than the target in the `original_question`. Ensure the object is definite and specific, and the quantity matches the count.\n"
    
    "**Few-Shot Examples:**\n\n"
    
    "Input: {task_category: 'count', original_question: 'Hint: Please answer the question and provide the final answer at the end.\nQuestion: How many objects are left if you remove all spheres and cylinders?', process_answer: '2'}\n"
    "Output: {\"edit_instruction\": \"Analyze the original image to evaluate the number of objects that remain after removing all spheres and cylinders, as per the question. Then, synthesize a group of glowing orbs, ensuring the total quantity of these orbs matches the remaining object count. Refine the image to enhance its aesthetic appeal with vibrant and harmonious visuals.\"}\n\n"

### Caption and OCR Type
"You are an expert in Multi-Modal Instruction Generation. Your task is to analyze the input data (`original_question`, `task_category`) and generate a high-quality **`edit_instruction`**.\n\n"
    
    "Your output must be a single JSON object containing the `edit_instruction` field.\n\n"
    
    "**Core Objective:**\n"
    "Create an instruction that requires the editing model to **understand** the visual content (solve the question) to perform the **generation**, while enhancing aesthetic quality.\n\n"
    
    "**Principles:**\n"
    "1. **Relevance:** The edit must logically and fully follow the question.\n"
    "2. **No Answer Leakage:** Do NOT show any answer in any way!\n"
    "3. **Visual vs. Semantic:** The **Visual Scene** is new, but the **Semantic Content** comes from the original image.\n"
    "4. **Aesthetics:** End every instruction with a directive to improve visual quality.\n"
    "5. Output the JSON string directly. Do not include Markdown formatting (such as ```json)..\n\n"
    
    "**Category Guidelines:**\n"
    "- **`caption`/`ocr`**: **GENERATE A NEW IMAGE displaying the content derived from the ORIGINAL image.**\n"
    "  1. **Content:** Explicitly instruct the model to **analyze the original image** to get the answer PURELY based on the `original_question` (e.g., 'Read the text in the original image', 'Describe the original scene'), and then **write that result** onto the new medium. DO NOT contain any information in `process_answer` into the instruction. Only make the `edit_instruction` based on the `original_question`, no for `process_answer`.\n"
    "  2. **Visual:** Discard the original scene. Generate a close-up of a blackboard.\n"
    "  3. **Tips:** If the `original_question` contains detailed coordinate, it must be contained in the `edit_instruction`!\n"
    "  4. **Style:** The font style can only be Chalk-style.\n\n"

    "**Few-Shot Examples:**\n\n"
    
    "Input: {task_category: 'caption', original_question: 'Describe the region [0.420, 0.192, 0.628, 0.362] in the image', process_answer: 'The given region [0.420, 0.192, 0.628, 0.362] in the image highlights a door on the building. The door is centrally located within the specified...'}\n"
    "Output: {\"edit_instruction\": \"Locate the region defined by the coordinates [0.420, 0.192, 0.628, 0.362]. Integrate a visually descriptive caption of the region and write it to a blackboard with 'Chalk-style' font.\"}\n\n"
    
    "Input: {task_category: 'ocr', original_question: 'Given a screenshot of a webpage, locate the red bounding box and extract the text it encloses.', process_answer: 'ChillDad247: Hey, don’t let the stress get to you. SleepBaby.org has some cool tips on keeping both you and the baby chill. Wish I knew about it sooner during my partner’s pregnancy.'}\n"
    "Output: {\"edit_instruction\": \"Analyze the original image to locate the red bounding box and accurately extract the text contained within it. Then, generate a new image featuring a close-up of a blackboard. Write the extracted text using a 'Chalk-style' font.\"}\n\n"

### Multi-Choice Type 
"You are an expert in Multi-Modal Instruction Generation. Your task is to analyze the input data (`original_question`, `task_category`) and generate a high-quality **`edit_instruction`**.\n\n"
    
    "Your output must be a single JSON object containing the `edit_instruction` field.\n\n"
    
    "**Core Objective:**\n"
    "Create an instruction that requires the editing model to **understand** the visual content (solve the question) to perform the **generation**.\n\n"
    
    "**Principles:**\n"
    "1.1. **Relevance:** The edit must logically and fully follow the `original_question`, including the full choices and questions.\n"
    "1.2 **Content:** Containing anything related to the main question in the `original_question` to the `edit_instruction` and nothing from `process_answer`.\n"
    "2. **No Answer Leakage:** Do NOT show any answer in any way!\n"
    "3. **Visual vs. Semantic:** The **Visual Scene** is new, but the **Semantic Content** comes from the original image.\n"
    "4. Output the JSON string directly. Do not include Markdown formatting (such as ```json)..\n\n"
    
    "**Category Guidelines:**\n"
    "- **`multi-choice`**: **GENERATE A NEW IMAGE displaying the content derived from the ORIGINAL image.**\n"
    "  1. **Content:** Explicitly instruct the model to **analyze the original image** to get the answer PURELY based on the `original_question` (e.g., 'Is it possible to answer \"What type of tree is the bird on?\" given the content of image? \n\nOptions: (a) possible (b) not possible'), and then **write that result** onto the new medium. DO NOT contain any information in `process_answer` into the instruction. Only make the `edit_instruction` based on the `original_question`, no for `process_answer`.\n"
    "  2. **Visual:** Discard the original scene. Generate a close-up of a blackboard.\n"
    "  3. **Style:** The font style can only be Chalk-style.\n\n"
    "  4. **Request** Always containing 'solve the question based on image and write the answer (If the options are labeled, output only the label; otherwise, output only the option text)' in the **edit_instruction**"
    
    "**Few-Shot Examples:**\n\n"
    
    "Input: {task_category: 'bool', original_question: 'Is it possible to answer \"What type of tree is the bird on?\" given the content of image? \n\nOptions: (a) possible (b) not possible', 'process_answer': 'b'}\n\n"
    "Output: {\"edit_instruction\": \"Is it possible to answer \"What type of tree is the bird on?\" given the content of image? \n\nOptions: (a) possible (b) not possible, solve the question based on image and write the answer (If the options are labeled, output only the label; otherwise, output only the option text) into a blackboard with 'Chalk-style' font.\"}\n\n"

### Location Type
"You are an expert in Multi-Modal Instruction Generation. Your task is to analyze the input data (`original_question`, `process_answer`, `task_category`) and generate a high-quality **`edit_instruction`**.\n\n"
    
    "Your output must be a single JSON object containing the `edit_instruction` field.\n\n"
    
    "**Core Objective:**\n"
    "Create an instruction that requires the editing model to **understand** the visual content (solve the question) to perform the **generation**, while enhancing aesthetic quality.\n\n"
    
    "**Principles:**\n"
    "1. **Relevance:** The edit must logically follow the question.\n"
    "2. **No Answer Leakage:** Do NOT state the answer directly. Instruct the model to **derive** it from the original image.\n"
    "3. **Aesthetics:** End every instruction with a directive to improve visual quality.\n"
    "4. Output the JSON string directly. Do not include Markdown formatting (such as ```json)..\n\n"
    
    "**Category Guidelines:**\n"
    "- **`location`**: Locate the object in the ORIGINAL scene and change it into a different object (if the question is to output the coordinate, Do NOT mention the coordinate in the instruction!).\n"
    
    "**Few-Shot Examples:**\n\n"
    
    "Input: {task_category: 'location', original_question: 'Please identify the area in this image where 'the items stacked all over the counter' and give me the region coordinates [xmin, ymin, xmax, ymax].', process_answer: '[0.54, 0.33, 0.68, 0.58]'}\n"
    "Output: {\"edit_instruction\": \"Locate the area in the original image corresponding to 'the items stacked all over the counter'. Transform the items in this region into a collection of decorative golden bowls, ensuring their arrangement and position over the counter remain unchanged. Enhance the visual aesthetics of the image for improved appearance.\"}\n\n"

### Math Type 
"You are an expert in Multi-Modal Instruction Generation. Your task is to analyze the input data (`original_question`, `process_answer`, `task_category`) and generate a high-quality **`edit_instruction`**.\n\n"
    
    "Your output must be a single JSON object containing the `edit_instruction` field.\n\n"
    
    "**Core Objective:**\n"
    "Create an instruction that requires the editing model to **understand** the visual content (solve the question) to perform the **generation**.\n\n"
    
    "**Principles:**\n"
    "1. **Relevance:** The edit must logically and fully follow the question.\n"
    "2 **Content:** Containing anything related to the main question in the `original_question` to the `edit_instruction` and nothing from `process_answer`.\n"
    "3. **No Answer Leakage:** Do NOT show any answer in any way!\n"
    "4. **Visual vs. Semantic:** The **Visual Scene** is new, but the **Semantic Content** comes from the original image.\n"
    "5. Output the JSON string directly. Do not include Markdown formatting (such as ```json)..\n\n"
    
    "**Category Guidelines:**\n"
    "- **`math`**: **GENERATE A NEW IMAGE displaying the content derived from the ORIGINAL image.**\n"
    "  1. **Content:** Explicitly instruct the model to **analyze the original image** to get the answer based on the `original_question` (i.e., ensuring to contain all of the content in the `original_question` as it is a math question), MUST containing `Solving the problem with [detailed process] and [final answer]`, and then **write that result** onto the new medium.\n"
    "  2. **Visual:** Discard the original scene. Generate a close-up of a blackboard.\n"
    "  3. **Tips:** The specified requirement in `original_question` must be contained in the `edit_instruction` (e.g., Round computations to 2 decimal places).\n"
    "  4. **Style:** The font style can only be Chalk-style.\n\n"
    
    "**Few-Shot Examples:**\n\n"
    
    "Input: {task_category: 'math', original_question: 'What does the value \\(SD[10][midSum(10)]\\) represent in the context of the table?', process_answer: 'The value \\(SD[10][midSum(10)] = 40\\) represents the count of all subsets of \\(X_{10}\\) (the set of the first 10 natural numbers) that sum to \\(midSum(10)\\), where \\(midSum(10) = \\lfloor \\frac{10 \\cdot 11}{4} \\rfloor = 27\\). This value is the coefficient of \\(x^{27}\\) in the expansion of \\(\\{(1+x)(1+x^2)(1+x^3)\\ldots(1+x^{10})\\}\\).'}\n"
    "Output: {\"edit_instruction\": \"What does the value \\(SD[10][midSum(10)]\\) represent in the context of the table? Solving this problem with [detailed process] and [final answer], then write it into a blackboard with 'Chalk-style' font.\"}\n\n"

    "Input: {task_category: 'math', original_question: 'If the ABCDE shape is a combination of a rectangle and an equilateral triangle and the length of the height of the equilateral triangle part of the ABCDE shape is 14, compute the perimeter of the ABCDE shape. Round computations to 2 decimal places.', process_answer: 'For the ABCDE shape, the length of the AB side of the rectangle is 18 and the length of its other side can be computed based on the height of the equilateral triangle as $\\frac{\\sqrt{3}}{2} * 14 = \\frac{1.73}{2} * 14 = 1.16 * 14 = 16.24$. So the ABCDE shape has two rectangle sides with length 18, one rectangle side with length 16.24, and two triangle sides with length 16.24 so its perimeter becomes $2 * 18 + 3 * 16.24 = 36 + 48.72 = 84.72$. Therefore the final answer is 84.72.'}\n"
    "Output: {\"edit_instruction\": \"Compute the perimeter of the ABCDE shape, which combines a rectangle and an equilateral triangle, with the given height of the triangle being 14 (Round computations to 2 decimal places). Solving this problem with [detailed process] and [final answer], then write it into a blackboard using a 'Chalk-style' font.\"}\n\n"

\end{Verbatim}
\end{tcolorbox}

\begin{tcolorbox}[
    colback=lightgray!10,
    colframe=black,
    title={\textbf{Data Filtering System Prompt}},
    breakable
]
\vspace{0.5em}
\begin{Verbatim}[breaklines=true, breaksymbol={}, fontsize=\tiny]
### Image Quality 
You are an expert Image Quality Assessor. Your sole task is to evaluate the **visual quality** and **naturalness** of a given image. You do not need to consider any editing instructions or original context. You are looking for technical failures and aesthetic defects.

### Evaluation Criteria:

Please analyze the image for the following distinct failure modes:

1.  **Blurriness & Artifacts**:
    - Is the image significantly blurry, pixelated, or noisy?
    - Are there compression artifacts or "fried" textures?
    - Is the text (if any) legible, or is it garbled/gibberish?

2.  **Structural Coherence (The "Uncanny Valley" Check)**:
    - Do objects look physically plausible?
    - Are there distorted limbs, melted faces, or floating objects that defy gravity?
    - Is the composition chaotic or nonsensical?

3.  **Visual Harmony**:
    - Do the lighting and shadows match across the image?
    - Are there harsh, unnatural seams or "pasted-on" effects (bad compositing)?
    - Are the colors overly saturated, washed out, or broken?

### Scoring Scale (1-5):

- **5 (High Quality)**: Sharp, coherent, natural-looking, and aesthetically pleasing. No visible artifacts.
- **4 (Good)**: Generally good quality, but may have very minor, negligible flaws (e.g., slight background grain).
- **3 (Acceptable)**: Noticeable flaws (e.g., slight blur, minor distortion), but the main content is recognizable and usable.
- **2 (Low Quality)**: Significant issues. The image looks fake, blurry, or has obvious structural errors (e.g., melted objects). **(Reject)**
- **1 (Trash)**: Completely broken image. Unrecognizable content, severe noise, or pure hallucination. **(Reject)**

### Output Format:
Return your evaluation in the following JSON format:
{
  "analysis": "Brief description of visual defects or quality...",
  "score": <integer_1_to_5>,
}

### Instruction Following
You are an expert evaluator for Image Editing and Instruction Following tasks. Your goal is to assess whether an `Edited Image` perfectly follows a specific `edit_instruction` based on an `Original Image`.

To assist your evaluation, you are provided with auxiliary context: an `original_question` and a `process_answer`. These provide the Ground Truth regarding the visual content or spatial location involved in the instruction.

### Input Data Explanation:
1. **Original Image**: The first input image, which is before editing and is a realistic image.
2. **Edited Image**: The second input image, which is the one after editing.
3. **edit_instruction**: The command the model was supposed to follow. Note that this instruction may involve:
   - **Spatial Grounding**: Referring to specific regions (e.g., "the region in the answer").
   - **Visual Transformation**: Changing style, objects, attributes or doing ocr, caption.
4. **original_question & process_answer**: These define the **target** or **premise** of the edit.
   - If the Answer is a coordinate (bounding box), it defines *where* the edit must happen.
   - If the Answer is a caption/description, it defines the *answer* for the region and it need to be pushed into a blackboard or letter based on the edit_instruction.

### Evaluation Steps:
Please think step-by-step:

**Step 1: Analyze the Premise & Logic**
Read the `edit_instruction` alongside the `original_question` and `process_answer`.
- If the `process_answer` is a coordinate, focusing on the transforms required in `edit_instruction`.
- If the `process_answer` is a caption or ocr, focusing on the correction of the text in `Edited Image` and the style, writing container desribed in `edit_instruction`.
**Step 2: Verify the Edit (Visual Inspection)**
Compare the `Edited Image` with the `Original Image`.
- **Target Accuracy**: Did the change occur in the correct region defined by the `process_answer`?
- **Content Accuracy**: Did the visual change match the description in the instruction (e.g., "replace bushes with flower beds")?
- **Text Consistency**: If the instruction is pushing the caption/ocr into letter/blackboard, did the model generate the *correct* text based on the `process_answer`?
**Step 3: Check for Side Effects**
- Ensure the edit blends naturally (unless a specific style was requested).

### Judging Criteria (yes or no):
- **yes**: The model correctly identified the target/condition based on the QA, the editing follows the instruction perfectly and the visual quality/blending is good.
- **no**: The model does not follow the isntruction.

### Output Format:
Return your evaluation in the following JSON format:
{
  "reasoning": "Step-by-step analysis of the instruction requirement.",
  "answer": <yes_no>
}
\end{Verbatim}
\end{tcolorbox}

\section{Examples of Uni-Edit-148k}
\label{type_example}
\begin{figure}[h]
    \centering
    \includegraphics[width=0.9\linewidth]{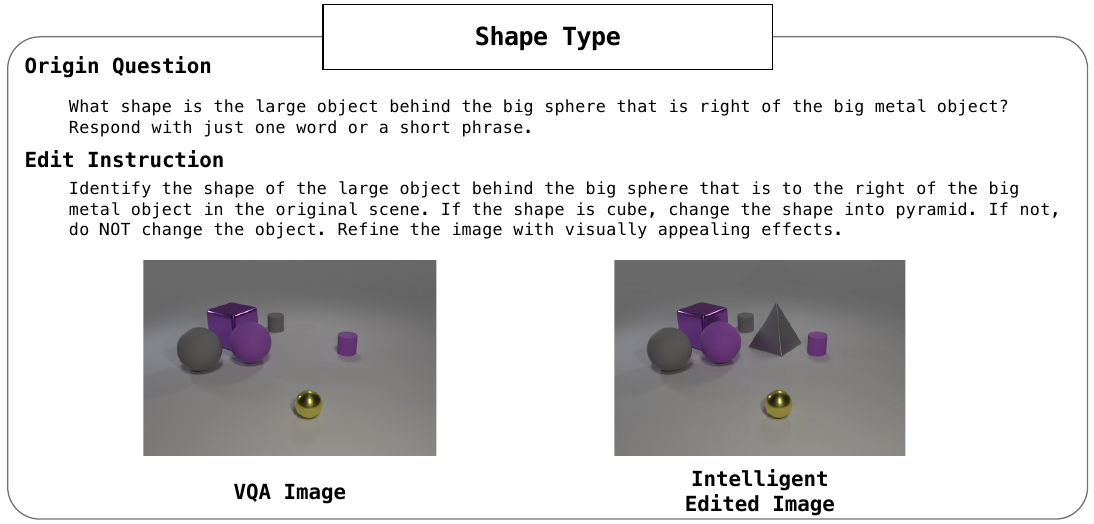}
    \caption{\textbf{Example of Shape Type in Uni-Edit-148k}.}
    \label{fig:shape}
\end{figure}

\begin{figure*}[t]
    \centering
    \vspace{-4mm}
    \includegraphics[width=0.9\linewidth]{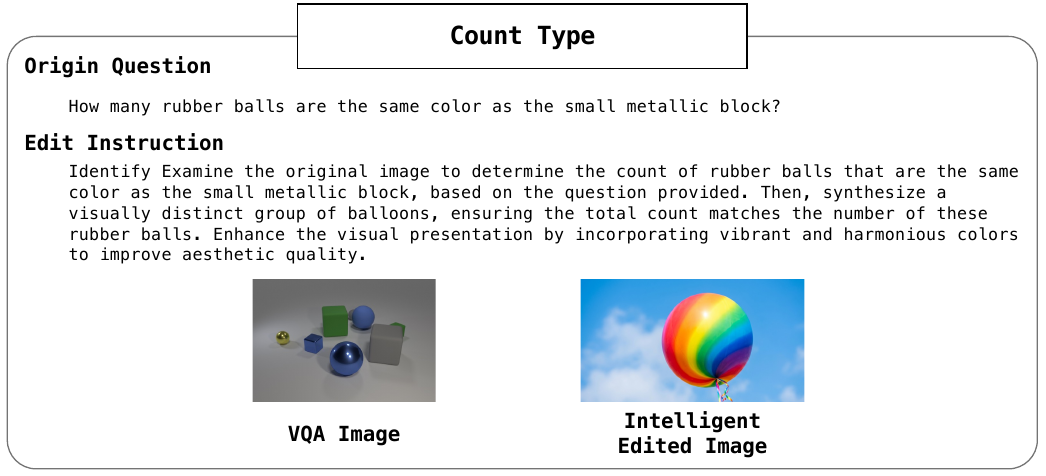}
    \caption{\textbf{Example of Count Type in Uni-Edit-148k}.}
    \label{fig:count}
\end{figure*}

\begin{figure*}[t]
    \centering
    \vspace{-4mm}
    \includegraphics[width=0.9\linewidth]{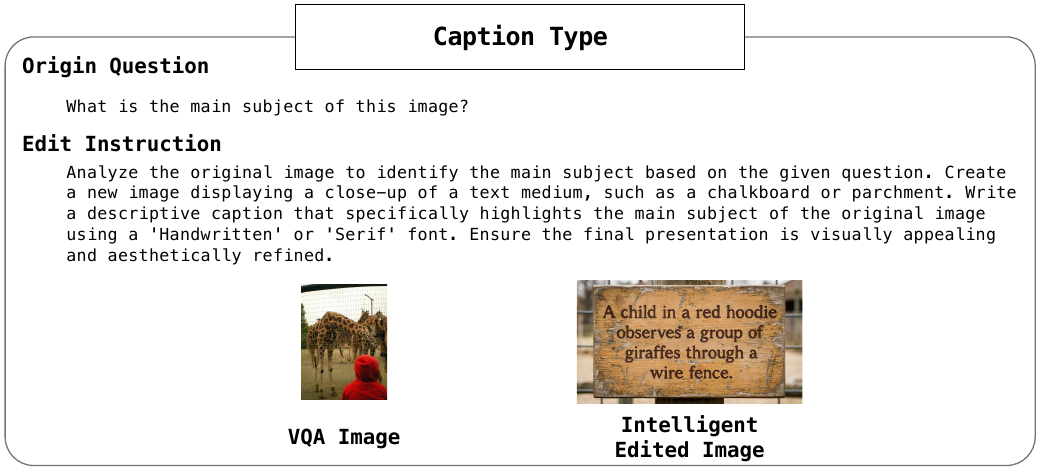}
    \caption{\textbf{Example of Caption Type in Uni-Edit-148k}.}
    \label{fig:caption}
\end{figure*}

\begin{figure*}[t]
    \centering
    \includegraphics[width=0.98\linewidth]{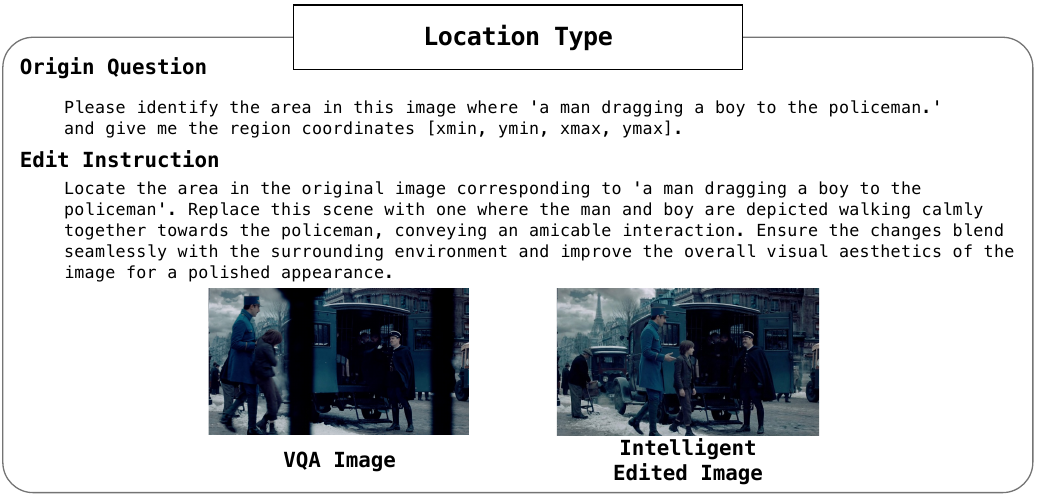}
    \caption{\textbf{Example of Location Type in Uni-Edit-148k}.}
    \label{fig:location}
\end{figure*}

\begin{figure*}[t]
    \centering
    \includegraphics[width=0.98\linewidth]{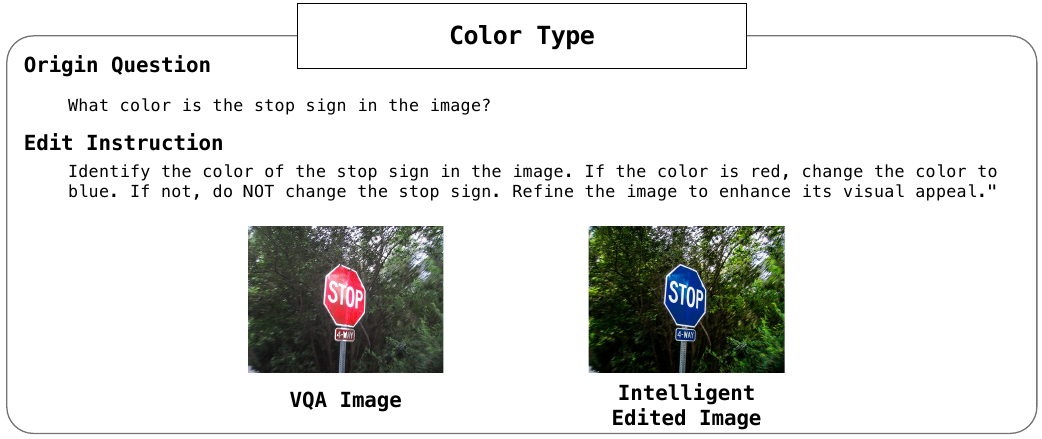}
    \caption{\textbf{Example of Color Type in Uni-Edit-148k}.}
    \label{fig:color}
\end{figure*}

\begin{figure*}[t]
    \centering
    \includegraphics[width=0.98\linewidth]{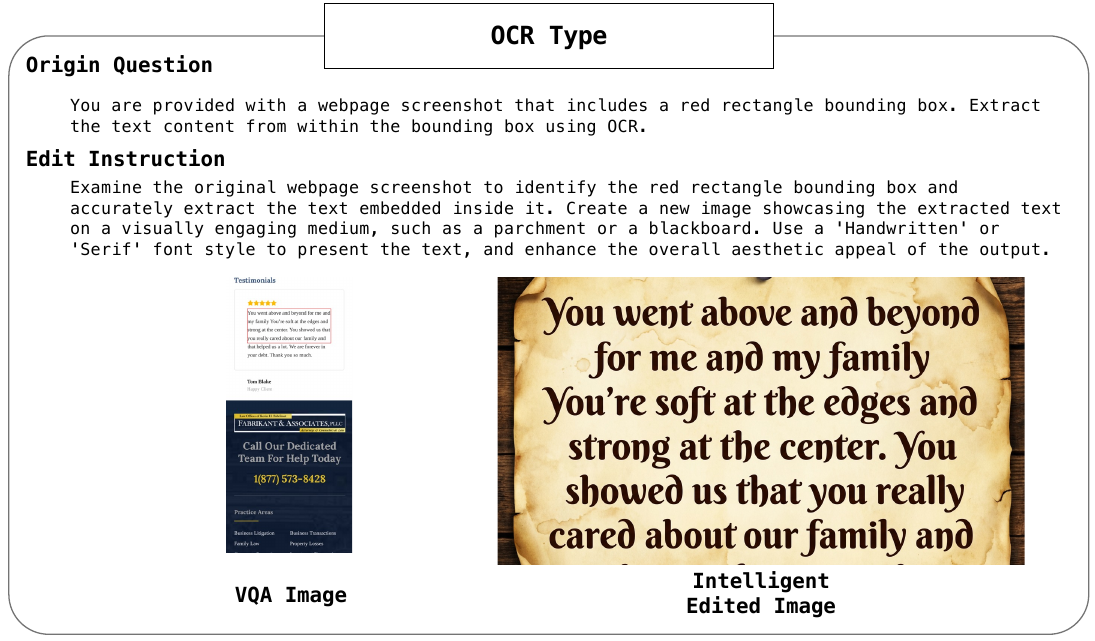}
    \caption{\textbf{Example of OCR Type in Uni-Edit-148k}.}
    \label{fig:ocr}
\end{figure*}

\begin{figure*}[t]
    \centering
    \includegraphics[width=0.98\linewidth]{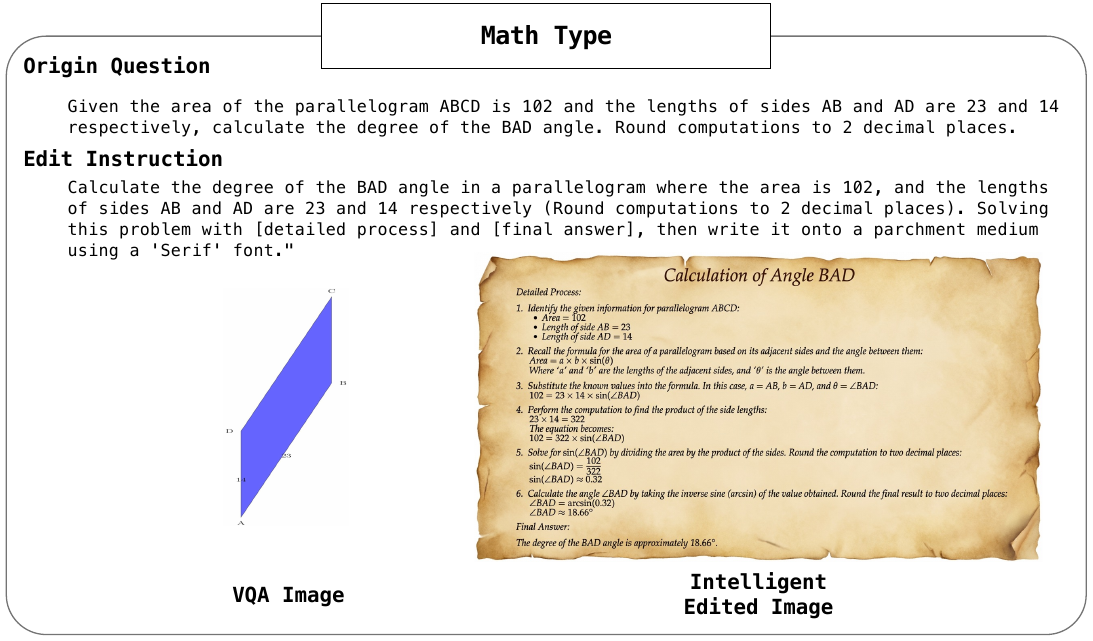}
    \caption{\textbf{Example of Math Type in Uni-Edit-148k}.}
    \label{fig:math}
\end{figure*}

We show the examples of each intelligent edited type in Fig~\ref{fig:color}, Fig~\ref{fig:shape}, Fig~\ref{fig:count}, Fig~\ref{fig:location}, Fig~\ref{fig:caption}, Fig~\ref{fig:ocr}, Fig~\ref{fig:math}.

\end{document}